\pdfoutput=1

\documentclass[11pt]{article}

\usepackage[]{acl}
\usepackage{times}
\usepackage{latexsym}
\usepackage{placeins}
\usepackage{booktabs}
\usepackage{natbib}  

\usepackage[T1]{fontenc}

\usepackage[utf8]{inputenc}
\usepackage{graphicx}
\usepackage{adjustbox}
\usepackage{multirow}
\usepackage{fontawesome}
\usepackage{amsmath}

\usepackage{microtype}

\usepackage{inconsolata}
\usepackage{CJKutf8}

\title{KG-TRICK~\faMagic: Unifying \underline{T}extual and \underline{R}elational \underline{I}nformation \underline{C}ompletion\\of \underline{K}nowledge for Multilingual Knowledge Graphs}

\author{
  Zelin Zhou \\ Apple \\ \texttt{charlie\_zhou@apple.com} \And{}
  Simone Conia \\ Sapienza University of Rome \\ \texttt{simone.conia@uniroma1.it} \And{}
  Daniel Lee \\ Adobe \\ \texttt{dlee1@adobe.com} \AND{}
  Min Li \\ Apple \\ \texttt{min\_li6@apple.com} \And{}
  Shenglei Huang \\ Apple \\ \texttt{huang\_kelsey@apple.com} \And{}
  Umar Farooq Minhas \\ Apple \\ \texttt{ufminhas@apple.com} \AND{}
  Saloni Potdar \\ Apple \\ \texttt{s\_potdar@apple.com} \And{}
  Henry Xiao \\ Apple \\ \texttt{henry\_xiao@apple.com} \And{}
  Yunyao Li \\ Adobe \\ \texttt{yunyaol@adobe.com}}

\newcommand{\secref}[1]{Section~\ref{#1}}

\newcommand{\tabref}[1]{Table~\ref{#1}}

\newcommand{\benchmark}{WikiKGE-10++}

\begin{document}
\maketitle

\begin{abstract}
  Multilingual knowledge graphs (KGs) provide high-quality relational and textual information for various NLP applications, but they are often incomplete, especially in non-English languages.
  Previous research has shown that combining information from KGs in different languages aids either Knowledge Graph Completion (KGC), the task of predicting missing relations between entities, or Knowledge Graph Enhancement (KGE), the task of predicting missing textual information for entities.
  Although previous efforts have considered KGC and KGE as independent tasks, we hypothesize that they are interdependent and mutually beneficial.
  To this end, we introduce KG-TRICK, a novel sequence-to-sequence framework that unifies the tasks of textual and relational information completion for multilingual KGs.
  KG-TRICK demonstrates that: i) it is possible to unify the tasks of KGC and KGE into a single framework, and ii) combining textual information from multiple languages is beneficial to improve the completeness of a KG.\@
  As part of our contributions, we also introduce \benchmark{}, the largest manually-curated benchmark for textual information completion of KGs, which features over 25,000 entities across 10 diverse languages.
\end{abstract}

\section{Introduction}\label{sec:introduction}
Knowledge graphs (KGs) aim to encode structured information about the world in a machine-readable format~\cite{hogan-etal-2021-knowledge}, providing high-quality relational and textual information for various NLP applications, such as question answering~\cite{mckenna-sen-2023-kgqa}, information retrieval~\cite{reinanda-etal-2020-knowledge}, entity linking~\cite{hu-etal-2023-entity}, and machine translation~\cite{modrzejewski-etal-2020-incorporating,conia-etal-2024-towards}, among others.
Although large language models (LLMs) are increasingly retrieving information from KGs to improve their factuality and performance in many NLP tasks~\cite{wang2023survey}, their effectiveness in multilingual applications is limited due to the important gap between the completeness of English and non-English information in KGs~\cite{peng-etal-2023-knowledge}.
In fact, KGs are not complete: a non-negligible quantity of information about entities (e.g., entity names, aliases, and descriptions) and relations (e.g., the connections between entities) is missing in non-English languages~\cite{conia-etal-2023-increasing}.
Therefore, improving the completeness of KGs has attracted significant attention over the years.

To address this issue, the research community has worked on two main tasks: Knowledge Graph Completion (KGC) and Knowledge Graph Enhancement (KGE).
KGC is the task of predicting missing relations between entities already defined in a KG~\cite{NIPS2013_1cecc7a7}, while KGE is the task of predicting missing textual information for entities in a KG~\cite{conia-etal-2023-increasing}.
More formally, KGC -- also known as link prediction -- is often defined as follows: given a KG $\mathcal{G}$, the task is to predict the missing tail entity $t$ given the head entity $h$ and the relation $r$ in a triplet $(h, r, ?)$.
For example, given the triplet \verb|(Joe Biden, occupation, ?)|, a possible answer could be \textit{politician} or, more specifically, the ID \verb|Q82955| of the \textit{politician} entity.\@
On the other hand, KGE is defined as follows: given an entity $e$ in a KG $\mathcal{G}$, the task is to predict missing textual information (e.g., an entity name, alias, or description) for $e$ in a target language.
For example, an alias for the entity \textit{Joe Biden} in English is \textit{Joseph R. Biden Jr.} or \textit{Joseph Robinette Biden Jr.}, while  its primary name in Chinese is \begin{CJK}{UTF8}{gbsn}\textit{乔·拜登}\end{CJK}. 
In this simple example, we can already see that there is an interdependence between KGC and KGE:\@ one entity can have different names in different languages, but the relation between entities should hold across languages.
However, unveiling this interdependence becomes challenging when dealing with ambiguous entities (e.g., \textit{Paris} the city and \textit{Paris} the prince of Troy) and entities whose names are not directly translatable (e.g., \textit{The Matrix} in English and \begin{CJK}{UTF8}{gbsn}黑客帝国\end{CJK} (\textit{Hacker's Empire}) in Chinese). 


Although KGC and KGE have previously been considered to be independent tasks, in this work, we investigate their interdependence and hypothesize that they are mutually beneficial.
Our hypothesis is based on two symmetric observations.
First, solving KGC provides rich language-independent relational information about entities, which may aid KGE to generate higher quality textual information across languages.
Second, solving KGE provides rich language-dependent textual information about entities, which may aid KGC in aligning relations among entities with names and descriptions across languages more effectively.

To this end, we introduce KG-TRICK (\underline{T}extual and \underline{R}elational \underline{I}nformation \underline{C}ompletion of \underline{K}nowledge), a novel unified framework that combines the tasks of KGC and KGE into a single task.
Different from previous approaches, the KG-TRICK framework is multilingual by design and is able to leverage the complementary textual information from multiple languages to improve the completeness of a multilingual KG.\@
Not only does KG-TRICK remove the need for separate KGC and KGE tasks, but it also outperforms similarly-sized state-of-the-art approaches tailored to each individual task, while achieving competitive performance compared to much larger language models.
To evaluate the robustness of KG-TRICK and encourage future systems on textual information completion of KGs, we also introduce \benchmark{}, the largest manually-curated benchmark for textual information completion for multilingual KG in 10 languages.

We can summarize our contributions as follows:
\begin{itemize}
  \item We unify the tasks of KGC and KGE to encompass not only the task of predicting missing links in a KG but also the task of completing its multilingual text;
  \item We introduce \benchmark{}, the largest manually-curated benchmark for textual information completion of KGs, including over 25,000 entities across 10 languages, to accompany KGC benchmarks and create a comprehensive evaluation suite;
  \item We present KG-TRICK, a novel sequence-to-sequence model that is able to combine information from multiple languages in an effective way to tackle textual and relation completion of knowledge graphs in a joint fashion;
  \item We show that KG-TRICK outperforms similarly-sized state-of-the-art models tailored for each task, while achieving competitive performance compared to larger LMs.
\end{itemize}

\noindent We believe that our work -- our task reformulation, manual benchmark, and unified method -- is a significant step forward to improve the quality of multilingual KGs and broaden their applicability to multilingual downstream tasks.
To encourage future work in this direction, we release \benchmark{} at \url{https://github.com/apple/ml-kge}.

\section{Related Work}\label{sec:related-work}
In this section, we briefly review the literature on Knowledge Graph Completion (KGC) and Knowledge Graph Enhancement (KGE) and discuss the challenges of completing textual and relational information in multilingual knowledge graphs.

\paragraph{Multilingual Knowledge Graphs.}
As mentioned above, KGs aim to encode information about our world knowledge in a structured, machine-readable format~\cite{hogan-etal-2021-knowledge}.
This information also includes lexicalizations, such as entity names, aliases, and descriptions;
when these are available in multiple languages, the KG is called a multilingual KG.\@
There are different ways to construct and organize multilingual KGs.
For example, in DBPedia~\cite{lehman-etal-2015-dbpedia}, an entity is language-dependent and is represented in different languages using different entity IDs.
Instead, in Wikidata~\cite{vrandevcic2014wikidata}, an entity is language-independent and is represented by the same entity ID to which different language-specific labels are attached.
The construction of multilingual KGs is an active area of research, and there are several challenges to be addressed, such as the alignment of entities across languages~\cite{chakrabarti-etal-2022-joint}, the completion of missing relational information~\cite{chen2020multilingual}, and the addition of textual information, especially in non-English languages~\cite{conia-etal-2023-increasing}.

\paragraph{Knowledge Graph Completion.}
The task of KGC is to predict missing relations between entities already defined in a KG~\cite{NIPS2013_1cecc7a7}. This task has been studied extensively in the literature, and there are several categories of methods, including embedding-based methods~\cite{lin-etal-2015-learning}, path-based methods~\cite{lin-etal-2015-modeling}, and rule-based methods~\cite{chen-etal-2020-knowledge}.
More recently, sequence-to-sequence models have been proposed to solve KGC by treating it as a text-to-text generation task, where the input is a partial triplet and the output is the missing entity~\cite{saxena-etal-2022-sequence}.
However, these approaches have been designed for monolingual KGs, as multilinguality adds a layer of complexity to the task.
More specifically, the completion of missing relations in a multilingual KG requires the ability to process and generate text in multiple languages.
\citet{chakrabarti-etal-2022-joint} have taken a step in this direction by including an auxiliary task to translate entity names, but they neither consider completing triples across languages nor the completion of more complex textual information, such as entity descriptions.

\paragraph{Knowledge Graph Enhancement.} The task of KGE is to predict missing textual information for entities in a KG.\@
This task is more recent in the literature, but there are several approaches to tackle it, such as machine translation, Web search, and language model-based methods~\cite{conia-etal-2023-increasing}.
However, \citet{conia-etal-2023-increasing} have mainly focused on i) combining answers from multiple KGE systems to improve coverage and precision, and ii) evaluating the quality of the textual information generated by KGE systems for popular entities only, while iii) ignoring the connection between KGE and KGC, especially in the multilingual setting.

\begin{figure*}[thbp] \centering
  \includegraphics[width=0.9\textwidth]{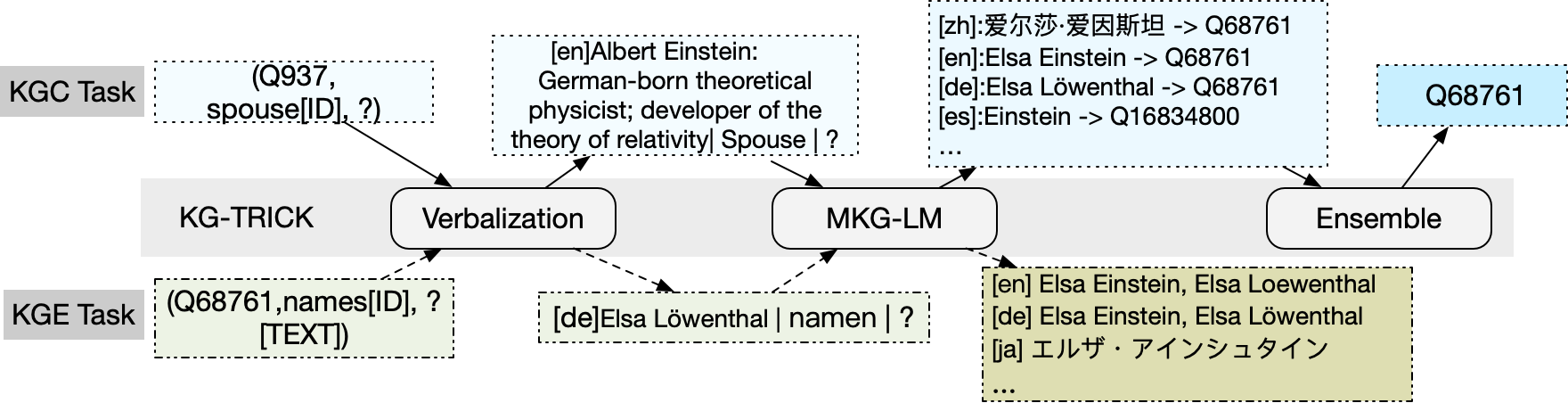}
  \caption{KG-TRICK: a unified seq-to-seq framework for KGC (dataflow in blue) and KGE (dataflow in green). For KGE, an input triplet $(\texttt{Q68761},\operatorname{names},?)$ is verbalized as ``[de] Elsa Löwenthal | namen | ?'' and then passed to the model, which generates outputs in multiple languages. For KGC, an input triplet $(\texttt{Q937},\operatorname{spouse},?)$ is verbalized as ``[en] Albert Einstein | spouse | ?'' and then passed to the model, which generates multilingual outputs and links to their corresponding IDs. The ensemble module consolidates all the outputs into the best one.}\label{fig:pipeline}
\end{figure*}

\section{Unifying Textual and Relational Information Completion}\label{sec:method}
In this section, we introduce KG-TRICK, or how we unify the tasks of KGC and KGE into a single framework, and how we leverage the complementary textual information
from multiple languages to improve the completeness of a multilingual KG.\@

\subsection{Task Reformulation}
Given the similarities between the two tasks of KGC and KGE and the interdependence between them (see \secref{sec:introduction}, in which we provide a high-level intuition), we reformulate both tasks as a single multilingual text-to-text generation task as shown in Figure~\ref{fig:pipeline}. KG-TRICK consists of three main components, namely the verbalization, the fine-tuned multilingual sequence-to-sequence model and the ensemble module to obtain the predicted entities for KGC task.
This unified framework allows us to i) treat KGC and KGE as a single task, and ii) leverage the complementary textual information from multiple languages to better complete factual information and reversely to improve the latent, dense representation of the fine-tuned sequence-to-sequence model, leading to improved KGE performance.
Figure~\ref{fig:pipeline} illustrates the pipeline of KG-TRICK for both KGC and KGE. Thanks to our reformulation, KG-TRICK sees both tasks as the task of predicting the tail entity $t$ given the head entity $h$ and the relation $r$ in a triplet $(h, r, ?)$, as we will detail in the following sections.


\subsubsection{KGC as Text-to-Text Generation}
In KGC, the task is to predict the missing tail entity $t$ given the head entity $h$ and the relation $r$ in a triplet $(h, r, ?)$.
We first reformulate this task as a text-to-text generation task, where the input is a partial triplet composed of the primary name and short description of the head entity $h$ and the relation $r$.
The model is then asked to generate the missing tail, or, more precisely, the primary name and short description of the tail entity $t$.


One important drawback of this reformulation is that it does not take into account the input and output languages, which is crucial for multilingual KGs.
We overcome this limitation by extending the triplet to a tuple of five elements, which include the source and target languages, as shown in Figure~\ref{fig:pipeline}.
More specifically, the input to the model is now a tuple $(l_s, l_t, h, r, ?)$, where $l_s$ is the source language of the input, $l_t$ is the target language of the output, $h = \textrm{primary name} + \textrm{short description}$, and $r$ is the relation.
The model then predicts $t$, i.e., the primary name and short description of the tail entity in $l_t$.
For example, given the input $(\textrm{en}, \textrm{es}, h, r, ?)$, the model generates the primary name and short description of the entity \textit{político | persona involucrada en la política}; while given the input $(\textrm{en}, \textrm{zh}, h, r)$, the model generates the primary name and short description of the entity
\begin{CJK}{UTF8}{gbsn}
  \textit{政治家 | 从事政治活动的人}
\end{CJK}. This reformulation significantly increases the training data pairs by extending cross-lingual name based entity alignment to cross-lingual relation based entity alignment resulting in higher quality of entity alignment.\@

\subsubsection{KGE as Text-to-Text Generation}
In KGE, the task is to predict missing textual information for entities in a KG.\@
This task can also be reformulated as a text-to-text generation task, similar to KGC.\@
We can immediately see that the formulation outlined above for KGC can be directly applied to KGE, with the only difference being that the head entity $h$ may be represented only by its primary name in case we want to generate a short description for $h$ itself.
Moreover, we also allow the head entity $h$ and the tail entity $t$ to be the same entity, which allows us to generate aliases for an entity in a specific language.
For example, given the partial
triplet \textit{Joe Biden: President of the US | has name |}, the model predicts the primary name of the entity \textit{Joe Biden} in the target
language but it can also generate one or more aliases, such as \textit{Joseph
  R. Biden Jr.} or \textit{Joseph Robinette Biden Jr.} in English, or
\begin{CJK}
  {UTF8}{gbsn}
  \textit{乔·拜登} or \textit{乔·罗宾内特·拜登} in Chinese.
\end{CJK}
Interestingly, when this reformulation is used in its most simple form, i.e., by
using only the primary name of the head entity, it becomes equivalent to translation into the target language.
This is a powerful feature, as it allows us to generate missing textual information in any language in KG.

\subsection{The KG-TRICK Model}
Unifying KGE and KGC, we implement KG-TRICK as a general sequence-to-sequence model, which learns to generate both missing relational and textual missing information in a KG.
More formally, given a tuple $(l_s, l_t, h, r, ?)$, the model is asked to generate $t$ from the
source language $l_s$ to the target language $l_t$ conditioned on $h$ and $r$ as
following:
\begin{equation}
  o = \textrm{KG-TRICK}(l_s, l_t, h, r, ?)
\end{equation}
where $o$ is the output generated by the model.
In other words, $o$ is the primary name and short description of the tail entity in the target language, and KG-TRICK learns to estimate the probability of generating $o$ given the input $(l_s, l_t, h, r, ?)$.

KG-TRICK can be implemented using any sequence-to-sequence architecture, such
as transformer~\cite{vaswani2017attention} or recurrent neural network~\cite{rumelhart1985learning}. In practice, we conducted our experiments with one main architecture, i.e., multilingual BART, which is a transformer-based encoder-decoder model, and we found
that it performs well on the task, as shown in \secref{sec:exp}.
The model can be trained using a standard maximum likelihood estimation (MLE) objective, and it can be evaluated using standard metrics for text generation, such as BLEU~\cite{papineni2002bleu}, ROUGE~\cite{lin-2004-rouge}, and COMET~\cite{rei-etal-2020-comet}.
In this work, we study three main variants of KG-TRICK, which differ in the training data they use: i) $\mathsf{TRICK_{KGC}}$, which uses only the relational information of the KG, ii) $\mathsf{TRICK_{KGE}}$, which uses only the textual information of the KG, and iii) $\mathsf{TRICK_{KGC+KGE}}$, which uses both the relational and textual information of the KG.\@

\begin{table*}[t]
    \centering
    \adjustbox{max width=\textwidth}{%
    \begin{tabular}{lcccccccccccc}
    \toprule    
    
     & \textbf{\#Entity} & \textbf{AR} & \textbf{DE} & \textbf{EN} & \textbf{ES} & \textbf{FR} & \textbf{IT} & \textbf{JA} & \textbf{KO} & \textbf{TH} & \textbf{ZH} & \textbf{All} \\
    
    \midrule
    
    \multirow{4}{*}{\rotatebox[origin=c]{90}{\textit{\small names}}} & \ Head & 1,000 & 1,000 & 1,000 & 1,000 & 1,000 & 1,000 & 1,000 & 1,000 & 1,000 & 1,000 & 10,000 \\
    & \ Torso & 750 & 750 & 750 & 750 & 750 & 750 & 750 & 750 & 750 & 750 & ~~7,500 \\
    & \ Tail & 750 & 750 & 750 & 750 & 750 & 750 & 750 & 750 & 750 & 750 & ~~7,500 \\
    & \ Total & 2,500 & 2,500 & 2,500 & 2,500 & 2,500 & 2,500 & 2,500 & 2,500 & 2,500 & 2,500 & 25,000 \\
    \cmidrule(l{3pt}r{3pt}){1-13}
    \multirow{4}{*}{\rotatebox[origin=c]{90}{\textit{\small descriptions}}} & \ Head & 958 & 986 & 983 & 971 & 974 & 988 & 920 & 906 & 882 & 923 & ~~9,491 \\
    & \ Torso & 894 & 942 & 918 & 926 & 919 & 936 & 773 & 828 & 790 & 822 & ~~8,748 \\
    & \ Tail & 896 & 936 & 916 & 857 & 901 & 928 & 728 & 728 & 869 & 767 & ~~8,526 \\
    & \ Total & 2,748 & 2,864 & 2,817 & 2,754 & 2,794 & 2,852 & 2,421 & 2,462 & 2,541 & 2,512 & 26,765 \\

    \bottomrule
    \end{tabular}}
    \caption{Overview of the number of entities in \benchmark{}, which features 10 languages -- Arabic (AR), German (DE), English (EN), Spanish (ES), French (FR), Italian (IT), Japanese (JA), Thai (TH) and simplified Chinese (ZH).}
    \label{tab:benchmark-overview}
\end{table*}

\subsection{Inference for KGC and KGE}
Once the output is generated, it can be used to complete the KG in two ways.
The application to KGE is straightforward, as the output is the missing textual information for an entity in the target language.
The application to KGC is slightly more complex, as the output is the textual representation (i.e., the primary name and short description) of the missing tail entity in the target language.
Relying only on an exact match between primary names may not be sufficient to determine the correct entity, especially in the case of ambiguous entities, such as \textit{Paris} the city and \textit{Paris} the prince of Troy.
While previous work~\cite{saxena-etal-2022-sequence} enumerates the entities with the same name (e.g., \textit{Paris$_1$}, \textit{Paris$_2$}, etc.), we incorporate entity descriptions as additional information from KG-TRICK to help disambiguate entities with the same primary name resulting in higher entity linking accuracy.


\paragraph{Ensemble across languages.}
Since KG-TRICK can generate text in any target language for which it has been (pre-)trained,
we leverage this capability to complete the missing information from multiple languages. When corresponding entity IDs are linked from generated text by different languages, we then ensemble and choose the most common predictions as is showed in Figure~\ref{fig:pipeline}.

\section{\benchmark{}}
\label{sec:benchmark}
In this section, we introduce \benchmark{}, the largest human-curated benchmark through expert crowdsourcing for textual information completion of KGs, which features over 25,000 entities and their corresponding aliases and descriptions across 10 languages.
Having realized the importance of evaluating systems on textual information completion of multilingual KGs, \citeauthor{conia-etal-2023-increasing} created WikiKGE-10, a benchmark for evaluating KGE systems on the completion of entity names and aliases in 10 languages.
However, WikiKGE-10 is limited in two dimensions: i) it only allows for the evaluation of entity names and aliases, and ii) the entities included in the benchmark are popular entities only (i.e., they belong to the top-10\% most popular entities in Wikidata according to number of page views of their corresponding Wikipedia pages).
A core contribution of our work is the creation of \benchmark{}, which extends WikiKGE-10 in two above-mentioned dimensions, hence the two ``+'' signs in the name of our benchmark. 
We include the number of entities across different languages and popularity tiers in \tabref{tab:benchmark-overview}.
The human annotation process, details and resources can be found in Appendix \ref{sec:benchmark-creation}.

\paragraph{Including entity descriptions.} \benchmark{} includes not only entity names and aliases, but also entity descriptions, which are crucial for many downstream tasks to create better entity representations~\cite{ri-etal-2022-mluke}.
The inclusion of entity descriptions in \benchmark{} is important for evaluating KGE systems on the completion of textual information that is usually longer and more complex than entity names and aliases.
However, evaluating KGE systems on the entity descriptions already available in Wikidata is not ideal, as those are not always manually curated and often under-specific, e.g., the description of many cities is simply \textit{city in country}.
Therefore, we asked a team of annotators to produce high-quality descriptions for each language. 

\paragraph{Including torso and tail entities.} \benchmark{} includes a much larger set of entities, which belong to the torso of the popularity distribution of Wikidata (i.e., between the top-10\% and top-50\% most popular entities) and also the tail of the popularity distribution (i.e., below the top-50\% most popular entities).
This is important as the majority of entities in a KG are not popular, and different conclusions can be drawn when evaluating systems on different popularity tiers.
The inclusion of torso and tail entities in \benchmark{} is important to assess the robustness of KGE systems on the completion of textual information for entities where the amount of information available inside (and also outside) the KG is limited.

Overall, \benchmark{} is around 2.5 times larger than WikiKGE-10 in terms of number of entities, while also including entity descriptions in addition to the entity names.

\section{Experiments and Results}\label{sec:exp}

\subsection{Datasets and Benchmarks}

\paragraph{KGC.} We carry out our KGC experiments on Wikidata5M~\cite{wang-etal-2021-kepler} transductive split.
We stress that, although entities in Wikidata are language-agnostic, the original Wikidata5M dataset is English-only, i.e., the textual information (names and descriptions) for the 5 million entities is only in English.
Therefore, this English-only setting may not be ideal to evaluate our multilingual KGC system; however, as is illustrated in \tabref{tab:kge_wikikge10v1}, 
our task reformulation could further enhance KGC performance with multilinguality.

\paragraph{KGE.} We carry out our KGE experiments on our newly annotated \benchmark{} dataset, which features 10 languages and over 25,000 entities as introduced in Section~\ref{sec:benchmark}.
We use this dataset to evaluate KGE models on the completion of entity names, aliases, and descriptions in 10 languages.

\subsection{Comparison Systems}

\paragraph{KGC.}
Baseline approaches for KGC can be divided into two categories: \textit{embedding-based} and \textit{text-based}.
Embedding-based methods derive an embedding for each entity and relation from the graph structure of the KG, and rank the most probable tail entity via a vector similarity function, e.g., L2 distance~\citep{NIPS2013_1cecc7a7}, complex space distance~\citep{trouillon2016complex} or other distance measures.
Text-based methods use encoder-only language models~\citep[SimKGC]{wang-etal-2022-simkgc} to encode both the head entity and the relation using their textual information, or encoder-decoder language models~\citep[KG-T5]{saxena-etal-2022-sequence} to generate the missing tail entity.
Recent work~\citep[KGT5-context]{kochsiek-etal-2023-friendly} integrates extra subgraph-structure information into sequence-to-sequence models. 
Since we build upon text-based methods with multilinguality, we compare our KG-TRICK models with SimKGC, KG-T5 and SKG-KGC~\citep{shan-etal-2024-multi}, which are the most relevant baselines for a fair comparison.

\paragraph{KGE.}
While KGE is a relatively recent task,
previous work has already indicated several strong baselines, including i) using NLLB-200\footnote{In this work we use NLLB-200-Distilled (600M).}~\cite{costa2022no} to translate entity names and descriptions from a language 
, and ii) prompting LLMs (e.g. GPT-3.5 or Llama), to generate textual information for an entity.

\subsection{Experimental Setup}
We use the entities within Wikidata5M which contains a set of around 5 million entities and a collection of around 20 million triplets and collect their available textual information (entity names, aliases, and descriptions) for English and 9 other languages from a Wikidata dump\footnote{Downloaded in November 2023.} to create a silver training set $\mathcal{E}$ for KGE in multiple EN$\rightarrow$XX directions containing around 16 million records.
For KGC, the original Wikidata5M triplets are expanded with our downloaded Wikidata dump to form over 150 million triplets $\mathcal{T}$ multilingually (EN$\rightarrow$XX) in 9 languages pairs.
We train three variants of KG-TRICK:\@ one on $\mathcal{E}$ (denoted as $\mathsf{TRICK_{KGE}}$), one on $\mathcal{T}$ (denoted as $\mathsf{TRICK_{KGC}}$), and one on the mixture of both (denoted as $\mathsf{TRICK_{KGC+KGE}}$). \footnote{All TRICK models are fine-tuned from mBART-large-50.}
We took care to prevent any test-set contamination by excluding test entity IDs from training within the WikiKGE10++ dataset. 
To verify the multilinguality of $\mathsf{TRICK_{KGE}}$, we also include a bilingual version trained with single language pair (e.g. EN$\rightarrow$IT) on KGE task, denoted as $\mathsf{TRICK_{KGE} (bilingual)}$ in \tabref{tab:kge_wikikge10v1}.
While the number of training samples are disproportional for KGC and KGE, to trade off the performance between the two tasks, as is shown in \tabref{tab:kgc_ratio}, we find a sweet spot of combining 50\% of KGC data into the joint training. We denote this derivative as $\mathsf{TRICK_{50\%KGC+KGE}}$ in \tabref{tab:kge_entity_popularity} and \tabref{tab:kge_wikikge10v1}. We provide more details balancing the training data of the two tasks in Appendix~\ref{sec:kgc_kge_ratio}.

\paragraph{Evaluation.}
For KGC, we evaluate the systems using standard ranking-based metrics, namely, \textit{hit@1}, \textit{hit@3}, \textit{hit@10}, and \textit{Mean Reciprocal Rank} (MRR).
Hit@k measures the proportion of correct answers in the top-k predictions, while MRR measures the average rank of the correct answer.
For KGE, we follow the evaluation protocol proposed by \citet{conia-etal-2023-increasing}, which includes two main metrics: coverage and precision.
\textit{Coverage} evaluates the number of entities for which a system is able to produce at least one correct entity name, while \textit{Precision} evaluates the ability of a system to identify incorrect entity names and aliases.
Finally, we report the COMET\footnote{We use the Unbabel/wmt22-cometkiwi-da variant.} scores for the completion of entity descriptions, a standard metric for text generation and machine translation. 

\subsection{Results on KGC}
Table \ref{tab:kgc_wikidata5m_transductive} shows the results obtained by our KG-TRICK models compared to other KGC-only models on the transductive test set of Wikidata5M.
In general, we can observe that KG-TRICK outperforms strong baselines on MRR, hit@1, and hit@3, and it is the third best model for hit@10.
More specifically, $\mathsf{TRICK_{KGC}}$ (trained only on KGC data but in multiple languages) already outperforms both SimKGC and KG-T5, on almost all the metrics.
Notably, this first result demonstrates that our model is able to outperform strong baselines that are tailored for KGC on a dataset Wikidata5M that is designed for KGC (and that is biased in its creation towards entities that have English lexicalizations).
Moreover, we can observe that $\mathsf{TRICK_{KGC+KGE}}$ achieves scores that
are even higher than $\mathsf{TRICK_{KGC}}$, which demonstrates that unifying KGC and KGE leads to additional improvements on the KGC task.
This second result empirically shows that the two tasks are indeed interdependent and mutually beneficial, and that the combination of the two tasks can lead to better results than the two tasks individually.


On the other hand, we can observe that the performance of $\mathsf{TRICK_{KGC}}$ and $\mathsf{TRICK_{KGC+KGE}}$ is not as good as the performance of SimKGC or SKG-KGC on hit@10.
We hypothesize that this is likely due to the fact that the sampling capacity of sequence-to-sequence models is limited and constrains their performance with higher values of $k$.
Indeed, for SimKGC, TransE and ComplEx, due to their closed-world assumption, the candidates for the tail entities are known during inference for similarity search. 
However, for text generation models, such as KG-T5 and KG-TRICK, which operate under a more challenging open-world assumption, 
the diversity of generated candidates may be limited by the sampling strategy used for decoding. 
We also include the results from KGT5-context paper which incorporates additional graph context into seq2seq models. 
Intuitively, adding graph context further alleviates the pain of text-based models capturing structured information within KGC task, which fulfills the shortcomings of generative methods in an orthogonal perspective than ours. 
Further work could focus on integrating graph context with multilinguality, and improving the sampling capacity of generative models.
\begin{table}[t]
  \adjustbox{max width=1.0\linewidth}{
    \setlength{\tabcolsep}{1.5pt}
    \begin{tabular}{lcccc}
      \toprule
      \textbf{Model}                        & \multicolumn{1}{l}{\textbf{MRR}} & \multicolumn{1}{c}{\textbf{hit@1}} & \multicolumn{1}{c}{\textbf{hit@3}} & \multicolumn{1}{c}{\textbf{hit@10}} \\
      \midrule
      TransE~\citep{NIPS2013_1cecc7a7}       & 25.3                             & 17.0                               & 31.1                               & 39.2                                \\
      DisMult~\citep{yang2014embedding}      & 25.3                             & 20.8                               & 27.8                               & 33.4                                \\
      SimpIE~\citep{kazemi2018simple}        & 29.6                             & 25.2                               & 31.7                               & 37.7                                \\
      RotatE~\citep{sun2018rotate}           & 29.0                             & 23.4                               & 32.2                               & 39.0                                \\
      QuatE~\citep{zhang2019quaternion}      & 27.6                             & 22.7                               & 30.1                               & 35.9                                \\
      ComplEx~\citep{trouillon2016complex}   & 30.8                             & 25.5.                              & -                                  & 39.8                                \\
      \midrule
      DKRL~\citep{Xie_Liu_Jia_Luan_Sun_2016} & 16.0                             & 12.0                               & 18.1                               & 22.9                                \\
      KEPLER~\citep{wang-etal-2021-kepler}   & 21.0                             & 17.3                               & 22.4                               & 27.7                                \\
      MLMLM~\citep{clouatre-etal-2021-mlmlm} & 22.3                             & 20.1                               & 23.2                               & 26.4                                \\
      SimKGC~\citep{wang-etal-2022-simkgc}   & 35.8                             & 31.3                               & 37.6                               & 44.1                       \\
      KG-T5~\citep{saxena-etal-2022-sequence}                               & 30.0                             & 26.7                               & 31.8                               & 36.5                                \\
      KG-T5 + Desc.~\citep{saxena-etal-2022-sequence}                        & 38.1                             & 35.7                               & 39.7                               & 42.2                                \\
      KG-T5 + Desc.$^{*}$                   & 37.0                             & 34.7                               & 38.4                               & 41.1\\
      SKG-KGC~\citep{shan-etal-2024-multi}                              & 36.6                             & 32.3                               & 38.2
         & \textbf{44.6} \\
      \midrule
      \textcolor[gray]{0.7}{KGT5-context~\citep{kochsiek-etal-2023-friendly}}     & \textcolor[gray]{0.7}{42.6}              & \textcolor[gray]{0.7}{40.6}              & \textcolor[gray]{0.7}{44.0}         & \textcolor[gray]{0.7}{46.0} \\
      \midrule
      $\mathsf{TRICK_{KGC}}$                & 38.2                             & 36.0                               & 39.7                               & 41.8                                \\
      $\mathsf{TRICK_{KGC+KGE}}$            & \textbf{38.8}                    & \textbf{36.6}                      & \textbf{40.4}                      & 42.6                                \\
      
      \bottomrule
    \end{tabular}
  }
  \caption{KGC results on the test set of Wikidata5M. $\mathsf{TRICK}$ achieves strong performance over competitive baselines. All cited scores are reported in their original paper. *: retrained in the same setting as $\mathsf{TRICK_{KGC}}$.}\label{tab:kgc_wikidata5m_transductive}
\end{table}

\begin{table*}[t]
    \centering
    \adjustbox{max width=1.0\textwidth}{
        \begin{tabular}{lccccccccccc}
            \toprule

                                                                     & \multicolumn{4}{c}{\textbf{Coverage}}
                                                                     & \multicolumn{4}{c}{\textbf{Precision}}
                                                                     & \multicolumn{3}{c}{\textbf{COMET}}
            \\

            \cmidrule(l{3pt}r{3pt}){2-5}\cmidrule(l{3pt}r{3pt}){6-9}\cmidrule(l{3pt}r{3pt}){10-12}

                                                                     & \textbf{Head}                          & \textbf{Torso} & \textbf{Tail} & \textbf{Avg}
                                                                     & \textbf{Head}                          & \textbf{Torso} & \textbf{Tail} & \textbf{Avg}
                                                                     & \textbf{Head}                          & \textbf{Torso} & \textbf{Tail}                                                                                                                                       \\

            \cmidrule(l{3pt}r{3pt}){1-12}

            NLLB-200$_{\small\ \texttt{EN} \rightarrow \texttt{XX}}$ & 29.1                                   & 26.1           & 24.3          & 26.5          & 47.6          & 39.6          & 34.9          & 40.7          & 0.64          & 0.63             & 0.63             \\

            Llama3-8B                                                & 27.2                                   & 22.9           & 20.4          & 23.5          & 46.0          & 36.6          & 31.2          & 37.9          & 0.62          & 0.62             & 0.62             \\
            Llama3-70B                                                & 31.6                                   & 26.8           & 24.1          & 27.5         & 49.0          & 39.9          & 34.7         & 41.2          & 0.60          & 0.60             & 0.62             \\

            GPT-3.5                                                  & \textbf{35.0}                          & 29.6           & 26.7          & 30.4          & 51.9          & 42.7          & 36.8          & 43.8          & \textbf{0.66} & \underline{0.64} & \underline{0.64} \\

            \cmidrule(l{3pt}r{3pt}){1-12}

            $\mathsf{TRICK_{KGE}}$                                   & 31.5                                   & \textbf{31.4}  & 29.9          & 30.9          & 56.4          & 51.4          & 46.5          & 51.4          & 0.63          & \underline{0.64} & \underline{0.64} \\
            $\mathsf{TRICK_{50\%KGC+KGE}}$                           & 33.1                                   & 31.2           & \textbf{30.1} & \textbf{31.5} & 57.7          & 51.7          & \textbf{47.1} & 52.1          & 0.61          & 0.62             & 0.62             \\
            $\mathsf{TRICK_{KGC+KGE}}$                               & 31.6                                   & 29.5           & 28.2          & 29.7          & \textbf{57.8} & \textbf{52.2} & 46.5          & \textbf{52.2} & 0.60          & 0.60             & 0.61             \\

            \bottomrule
        \end{tabular}
    }
    \caption{KGE results on WikiKGE-10++ split by head, torso and tail entities. Best results in bold.}
    \label{tab:kge_entity_popularity}
\end{table*}

\begin{table*}[h!]
  \centering
  \adjustbox{max width=1.0\textwidth}
  {
    \begin{tabular}{llcccccccccc}
      \toprule

      \multirow{17}{*}{\rotatebox[origin=c]{90}{\textit{names \& aliases}}} & \textbf{Coverage F1}                                     & \textbf{\#Params}     & \textbf{AR}      & \textbf{DE}      & \textbf{ES}      & \textbf{FR}      & \textbf{IT}   & \textbf{JA}   & \textbf{KO}   & \textbf{ZH}   & \textbf{Avg}  \\

      \cmidrule(l{3pt}r{3pt}){2-12}

                                                                            & NLLB-200$_{\small\ \texttt{EN} \rightarrow \texttt{XX}}$ & ~0.6B                 & 16.9             & 39.8             & 37.9             & 40.8             & 40.5          & 9.4           & 18.6          & 8.1           & 26.5          \\
                                                                            & Llama3-8B                                                & ~8B                   & 11.3             & 35.9             & 32.8             & 35.0             & 32.6          & 12.4          & 15.5          & 12.5          & 23.5          \\
                                                                            & Llama3-70B                                                & ~70B                   & 20.3             & 39.1             & 35.7             & 38.9             & 38.6          & 15.6          & 18.0          & 13.5          & 27.5          \\
                                                                            & GPT-3.5                                                  & 175B                  & 20.1             & 40.8             & 39.1             & 41.3             & 40.9          & 19.6          & 21.5          & \textbf{20.0} & 30.4          \\

      \cmidrule(l{3pt}r{3pt}){2-12}

                                                                            & $\mathsf{TRICK_{KGE} (bilingual)}$                       & \multirow{4}{*}{0.6B} & \underline{24.4} & 39.5             & 33.7             & 37.2             & 34.0          & 18.4          & 23.3          & 15.3          & 27.5          \\
                                                                            & $\mathsf{TRICK_{KGE}}$                                   &                       & \underline{24.4} & 39.5             & 38.8             & 40.8             & 40            & 20.4          & 26.3          & 17.0          & 30.9          \\
                                                                            & $\mathsf{TRICK_{50\%KGC+KGE}}$                           &                       & 23.0             & \textbf{40.9}    & \textbf{40.0}    & \textbf{41.9}    & \textbf{41.1} & \textbf{20.5} & \textbf{26.0} & 18.2          & \textbf{31.5} \\
                                                                            & $\mathsf{TRICK_{KGC+KGE}}$                               &                       & 22.6             & 39.2             & 37.2             & 39.4             & 39.4          & 19.8          & 24.3          & 16.0          & 29.7          \\

      \cmidrule(l{3pt}r{3pt}){2-12}
                                                                            & \textbf{Precision F1}                                    & \textbf{\#Params}     & \textbf{AR}      & \textbf{DE}      & \textbf{ES}      & \textbf{FR}      & \textbf{IT}   & \textbf{JA}   & \textbf{KO}   & \textbf{ZH}   & \textbf{Avg}  \\

      \cmidrule(l{3pt}r{3pt}){2-12}
                                                                            & NLLB-200$_{\small\ \texttt{EN} \rightarrow \texttt{XX}}$ & ~0.6B                 & 30.3             & 49.1             & 48.9             & 51.2             & 52.0          & 28.3          & 30.6          & 34.9          & 40.7          \\
                                                                            & Llama3-8B                                                & ~8B                   & 24.1             & 46.0             & 45.0             & 47.7             & 45.4          & 30.4          & 26.6          & 38.3          & 37.9          \\
                                                                            & Llama3-70B                                                & ~70B                   & 33.0             & 48.1             & 47.1             & 50.4             & 49.5          & 33.5          & 29.1         & 38.6          & 41.2          \\
                                                                            & GPT-3.5                                                  & 175B                  & 33.2             & 49.5             & 49.7             & 51.9             & 52.3          & 36.9          & 33.1          & 43.8          & 43.8          \\
      \cmidrule(l{3pt}r{3pt}){2-12}
                                                                            & $\mathsf{TRICK_{KGE} (bilingual)}$                       & \multirow{4}{*}{0.6B} & 46.3             & 52.3             & 53.1             & 54.5             & 55.3          & 43.8          & 46.3          & 45.5          & 49.6          \\
                                                                            & $\mathsf{TRICK_{KGE}}$                                   &                       & 48.4             & 54.3             & 55.2             & \textbf{57.0}    & 56.2          & 45.6          & 47.5          & 47.5          & 51.4          \\
                                                                            & $\mathsf{TRICK_{50\%KGC+KGE}}$                           &                       & 48.1             & \textbf{55.9}    & \textbf{55.6}    & 56.4             & \textbf{56.9} & 45.7          & \textbf{50.0} & \textbf{48.5} & 52.1          \\
                                                                            & $\mathsf{TRICK_{KGC+KGE}}$                               &                       & \textbf{49.4}    & 54.8             & 55.5             & 56.7             & 56.6          & \textbf{47}   & 49.0          & 48.3          & \textbf{52.2} \\

      \midrule
      \multirow{8}{*}{\rotatebox[origin=c]{90}{\textit{descriptions}}}      & \textbf{COMET Score}                                     & \textbf{\#Params}     & \textbf{AR}      & \textbf{DE}      & \textbf{ES}      & \textbf{FR}      & \textbf{IT}   & \textbf{JA}   & \textbf{KO}   & \textbf{ZH}   & \textbf{Avg}  \\

      \cmidrule(l{3pt}r{3pt}){2-12}
                                                                            & NLLB-200$_{\small\ \texttt{EN} \rightarrow \texttt{XX}}$ & ~0.6B                 & 0.59             & 0.62             & \underline{0.66} & 0.63 & 0.65          & 0.63          & 0.65          & 0.64          & 0.63          \\
                                                                            & Llama3-8B                                                & ~8B                   & 0.56             & 0.62             & 0.65             & 0.63 & 0.65          & 0.60          & 0.60          & 0.61          & 0.62          \\
                                                                            & Llama3-70B                                                & ~70B                   & 0.59             & \underline{0.63}             & \textbf{0.67}             & \textbf{0.64} & 0.66          & 0.56          & 0.56          & 0.54          & 0.60          \\
                                                                            & GPT-3.5                                                  & 175B                  & \textbf{0.60}    & \underline{0.63} & 0.66 & 0.63 & \textbf{0.66} & 0.66          & \textbf{0.67} & \textbf{0.67} & \textbf{0.65} \\
      \cmidrule(l{3pt}r{3pt}){2-12}
                                                                            & $\mathsf{TRICK_{KGE}}$                                   & \multirow{3}{*}{0.6B} & 0.58             & \underline{0.63} & 0.66 & \underline{0.63} & 0.65          & \textbf{0.67} & 0.64          & 0.64          & 0.64          \\
                                                                            & $\mathsf{TRICK_{50\%KGC+KGE}}$                           &                       & 0.56             & 0.61             & 0.63             & 0.60             & 0.64          & 0.64          & 0.62          & 0.60          & 0.61          \\
                                                                            & $\mathsf{TRICK_{KGC+KGE}}$                               &                       & 0.55             & 0.59             & 0.63             & 0.58             & 0.62          & 0.63          & 0.62          & 0.60          & 0.60          \\

      \bottomrule
    \end{tabular}
  }
  \caption{KGE experiments on WikiKGE-10++ split by languages. TRICK achieves best performance on Precision and Coverage F1 scores. Best results in bold.}
  \label{tab:kge_wikikge10v1}
\end{table*}

\subsection{Results on KGE}
Table~\ref{tab:kge_entity_popularity} shows the KGE results on our new \benchmark{} benchmark split by entity popularity, while Table~\ref{tab:kge_wikikge10v1} shows the results by language. 

\paragraph{Coverage.} 
Overall, we could observe that $\mathsf{TRICK_{50\%KGC+KGE}}$ outperforms strong baselines on average and across most languages. 
Interestingly in \tabref{tab:kge_entity_popularity}, $\mathsf{TRICK_{KGE}}$ and $\mathsf{TRICK_{KGC+KGE}}$ perform worse than GPT-3.5 on Coverage of head entities.
This is likely due to the fact that GPT-3.5 has seen substantially more popular entity names during its pre-training and is equipped with considerably more parameters to store such information.
However, $\mathsf{TRICK}$ series quickly catch up with GPT-3.5 on Coverage of torso entities, and significantly outperform GPT-3.5 on Coverage of tail entities.
It shows that GPT-3.5 quickly loses its advantage when the entities are less popular, and that KG-TRICK models feature a more balanced and consistent performance across different popularity tiers.

\paragraph{Precision.} Overall, we can observe that $\mathsf{TRICK}$ significantly outperforms
strong baselines on precision on head, torso, and tail entities, i.e., it
is a more reliable system in identifying incorrect entity names and aliases in
a given target language in a multilingual knowledge graph. In fact,
$\mathsf{TRICK}$ is particularly effective on torso and tail entities, where it
improves over NLLB-200 and GPT-3.5 by around 10\% points in F1 score.
This is important as completing missing knowledge is not only about providing
the correct information but also about avoiding incorrect information.

\paragraph{Entity descriptions.} Finally, we also report the COMET score for the completion of entity
descriptions, borrowing a metric for open-ended text generation from MT. 
In this task, we can observe that $\mathsf{TRICK_{KGE}}$ is comparable with
NLLB-200 and GPT-3.5, 
while $\mathsf{TRICK_{KGC+KGE}}$ is slightly worse on
average than $\mathsf{TRICK_{KGE}}$. These results open the door to future
work: indeed, very different methods achieve comparable results on entity
description completion, meaning that there is still a wide room for improvement
in this task or COMET is not a good metric for comparing descriptions. 
We note that BLEU is not appropriate either, as its score is not defined for short texts, e.g., one word.

\paragraph{Multilinguality and Multi-tasking.}
As is illustrated in \tabref{tab:kge_wikikge10v1}, $\mathsf{TRICK_{KGE}}$ outperforms $\mathsf{TRICK_{KGE} (bilingual)}$ in almost all languages both on Precision and Coverage, indicating that jointly training a unified model for all languages can inherently benefit its multilingual capabilities. On the multi-task side, combining KGC and KGE tasks requires caution, as is demonstrated by $\mathsf{TRICK_{KGC+KGE}}$ and $\mathsf{TRICK_{50\%KGC+KGE}}$. KGC and KGE are mutually beneficial when training data is balanced, otherwise one task dominates the distribution and causes a regression on the other. A more in-depth analysis focused on data balancing is discussed in Appendix~\ref{sec:kgc_kge_ratio}. We can also observe that the multilingual capability of Llama3 is far from ideal in KGE.

\section{Downstream Application: Results on Multilingual Question Answering}
\label{sec:KGQA}
In addition to the KGC and KGE tasks, we also evaluate our KG-TRICK on a
downstream application: our intuition is that (post-)pretraining on KGC and KGE
tasks can allow a model to store more factoid knowledge, which can be useful
for multilingual question answering. Therefore, we evaluate the performance of
our $\mathsf{TRICK_{KGC+KGE}}$ when fine-tuned on answering the questions in the Mintaka
dataset~\cite{sen-etal-2022-mintaka}, which is a multilingual question answering dataset that
contains knowledge-seeking questions, and compare its results in the same setting with directly fine-tuning on Mintaka using mBART-large-50 that has not been (post-)pretrained on KGC and KGE tasks.
Our experiments show that our model
outperforms mBART-large-50 by 3.1\% (29.2\% vs. 26.1\%) on average in terms of EM (Exact Match), which demonstrates
that the knowledge embedded in KGC and KGE training data could be easily
transferred to the QA task in cross-lingual settings.

\section{Conclusion}
The contributions of this paper are threefold. First, we propose a novel
multilingual KGC and KGE system, $\mathsf{TRICK}$, which is able to complete
relational and textual information in and across 10 languages. Second, we
introduce a largest human-curated dataset, \benchmark{}, which contains 10
languages and over 25,000 entities for KGE evaluation.  
Third,
we demonstrate that our $\mathsf{TRICK}$ system outperforms strong baselines on
both KGC and KGE tasks, 
and that the combination of the two tasks can lead to
better results than addressing them individually. In addition, we also show that the knowledge
embedded in KGC and KGE training data could be easily transferred to cross-lingual QA.
We hope our work and our \benchmark{} can inspire future research on multilingual KGC and KGE, and further contribute to both KG and LLM communities on evaluating the factuality of Language Models across languages.


\section*{Limitations}

\paragraph{Closed world assumption of KGC.} In this paper, we assume that the entities within KGC tasks already exist in the KG. If the model predicts some entities that do not exist, we simply ignore the inference. This assumption limits the model's capability to explore the encoded knowledge within pre-trained multilingual LMs.
Although our model can be extended to predict and extract entities outside of a KG, our experiments in~\secref{sec:exp} demonstrate that there is still a large headroom to complete the relational information in a multilingual setting, since we combine knowledge across languages that are not considered in a monolingual setting. We leave the exploration of how to leverage KG-TRICK to work in an open-world setting for future work.

\paragraph{Limited exploration of pre-trained multilingual LMs.} Our KGE task pays great attention to the enrichment of entity names and entity descriptions with limited attention to other textual information such as mottos, quotes, and others. Given that pre-trained LMs have been trained on massive amounts of data, there is great potential in the extraction of information that has been seen by the pre-trained LMs and that does not exist in the KG yet. Although KG-TRICK can be extended to infer other entity facts, our analysis shows that entity names and descriptions are still the most essential pieces of information to enrich and disambiguate entities, especially the entity descriptions, as they are a summary in free-form text that is highly representative of a particular entity.

\paragraph{Support unified multilingual KGs.} We focus on the multilingual KGs in which entities are represented by entity IDs, and language-dependent textual information is structured as attributes associated with the corresponding entities.
The benefit of this setting is that we enriched a KG that has been unified across languages at the very beginning of its construction process, and add new knowledge to the KG itself by inferring information that can be derived from the multilingual KG itself. Therefore, our system does not suffer from the error propagation introduced by entity and relation alignment between different KGs. Nonetheless, our system is limited to the setting of unified multilingual KGs, such as Wikidata. KG-TRICK is complementary to other related work on multilingual KG completion, which calls for integration of different KGs. Our system can be applied to further improve the completeness after the KGs are unified since such techniques focus on fusing different KGs but not inferring knowledge from the unified KG itself.

\paragraph{\benchmark{}.} While \benchmark{} significantly extends WikiKGE-10 by adding a significant number of entities sampled from torso and tail entities, it contains only two types of facts, i.e., entities names (and aliases) and descriptions. Future work may extend \benchmark{} to cover more types of facts that are usually associated with entities, so that the research community will be able to get a more accurate and thorough picture on how to evaluate novel approaches in this area.

\paragraph{Potential risks for generative textual information completion.} As we employ a text-to-text framework to complete the information in a multilingual KG, it may generate biased or inaccurate text that could be misleading for downstream tasks. If this work is considered for production use, human annotators should be added in the loop to reduce the risks of harmful text generation.

\section*{Acknowledgements}
We would like to thank all the people at Apple who provided their feedback on this work and participated in many helpful conversations. Part of this work was carried out when Simone Conia and Daniel Lee were interns at Apple. Simone Conia gratefully acknowledges the support of the PNRR MUR project PE0000013-FAIR, which fully funds his fellowship since October 2023.

\bibstyle{acl_natbib}
\bibliography{anthology,custom}

\newpage

\appendix

\section{Creating \benchmark{}}
\label{sec:benchmark-creation}

In this section, we describe the in-depth details on the creation of \benchmark{}, our novel human-curated dataset for the evaluation of automatic approaches on KGE of Wikidata entity names and descriptiptions.

\subsection{Choice of Languages}
Aligned with the previous work completed in \citeauthor{conia-etal-2023-increasing}, the benchmark, we select 9 languages from a set of typologically diverse linguistic families, while replacing the Russian (Slavic) language for the Thai (Kra–Dai) language:

\begin{itemize}
  \item West Germanic: English, German;
  \item Romance: Spanish, French, Italian;
  \item Semitic: Arabic;
  \item Sino-Tibetan: Chinese (simplified);
  \item Kra–Dai: Thai;
  \item Koreanic: Korean;
  \item Japonic: Japanese.
\end{itemize}

The Russian language was interchanged for the Thai language due to export and import restrictions placed on Russia, thereby, restricting access to Russia-based human annotators.

\subsection{Human annotation process}
The objective of the annotation process was to (i) rate and suggest entity names in the target language, (ii), verify the suggest entity names in the target languages, (iii) curate description for the entity in the target language, (iv) validate the provided descriptions quality.

\subsubsection{Rate and suggest entity names.}
The objective of this annotation step was to rate entity names in a target language. Detailed information on the annotation process and UI design can be found in \citeauthor{conia-etal-2023-increasing}.

\subsubsection{Verify suggested entity names.}
The objective of this annotation step was to verify the suggested entity names in a target language provided by the human annotators. Detailed information on the annotation process and UI design can be found in \citeauthor{conia-etal-2023-increasing}.

\subsubsection{Curate entity descriptions.}
The objective of this annotation step was to curate descriptions for a given entity in the target language.

Given an entity name in a target language, annotations were required to familiarize themselves with its information: the user interface provided the entity names, as well as a built-in panel that directly displayed Wikipedia articles for the corresponding entity in English and the target language, if available. In addition, annotators were recommended to further familiarize themselves with the entity outside of the provided information.

\begin{figure*}[t]
  \centering
  \includegraphics[width=\textwidth]{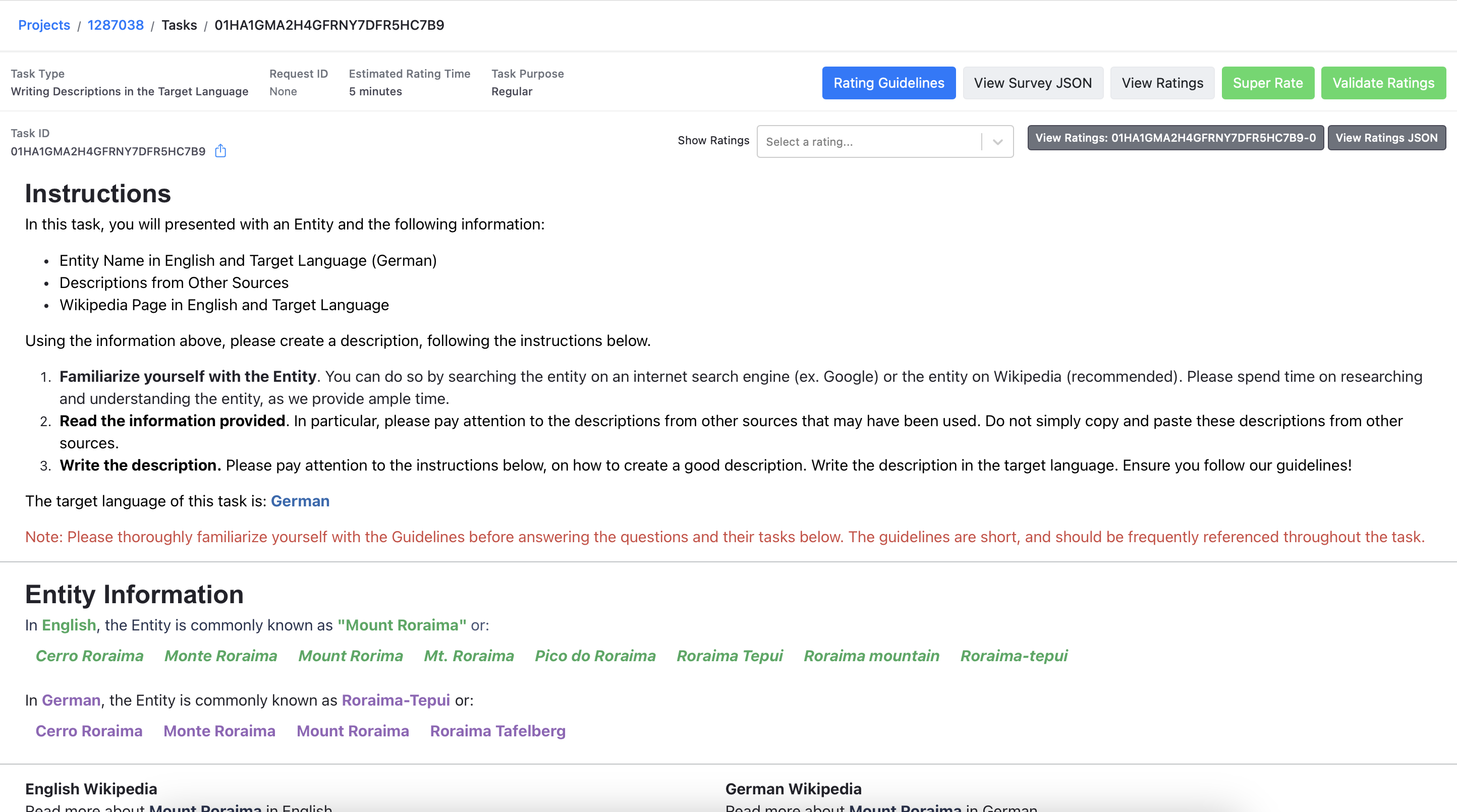}
  \caption{UI used for the annotation task: the annotators could familiarize themselves with the task with an outline of the task instructions (detailed guidelines could be read in a separate page) and the information about the entity, including its names in English and its Wikipedia pages in English and the target language (Italian in this case).}
  \label{fig:ui-first-part}
\end{figure*}

Next, the annotators were tasked with learning about the required format of the requested description with detailed instructions. This was facilitated by providing: (i) examples of correctly curated description given an example entity, and (ii) strict rules that the descriptions had to comply by.

\begin{figure*}[t]
  \centering
  \includegraphics[width=\textwidth]{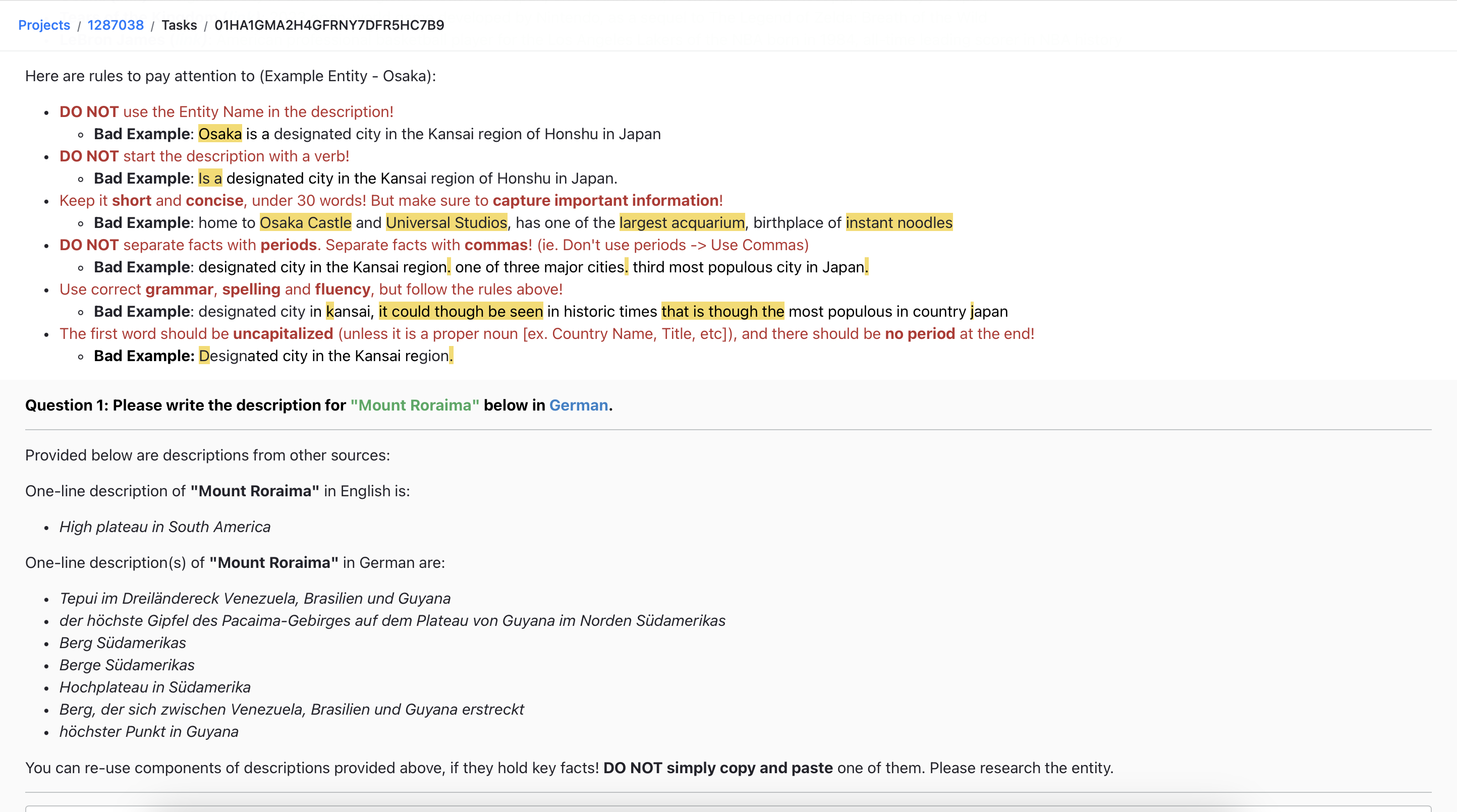}
  \caption{UI used for the annotation task: the annotators familiarized themselves with the description format with an outline of the requirements (detailed guidelines could be read in a separate page).}
  \label{fig:ui-second-part}
\end{figure*}

After learning about the entity and the required description format, the human annotator was requested to manually curate the description for the corresponding entity in the target language. During this task, the human annotator was provided with descriptions from other sources (such as Wikidata) in English and the target language. Human annotators were instructed that they could leverage the extraneous descriptions, but not to copy and paste unless satisfactory.

\begin{figure*}[t]
  \centering
  \includegraphics[width=\textwidth]{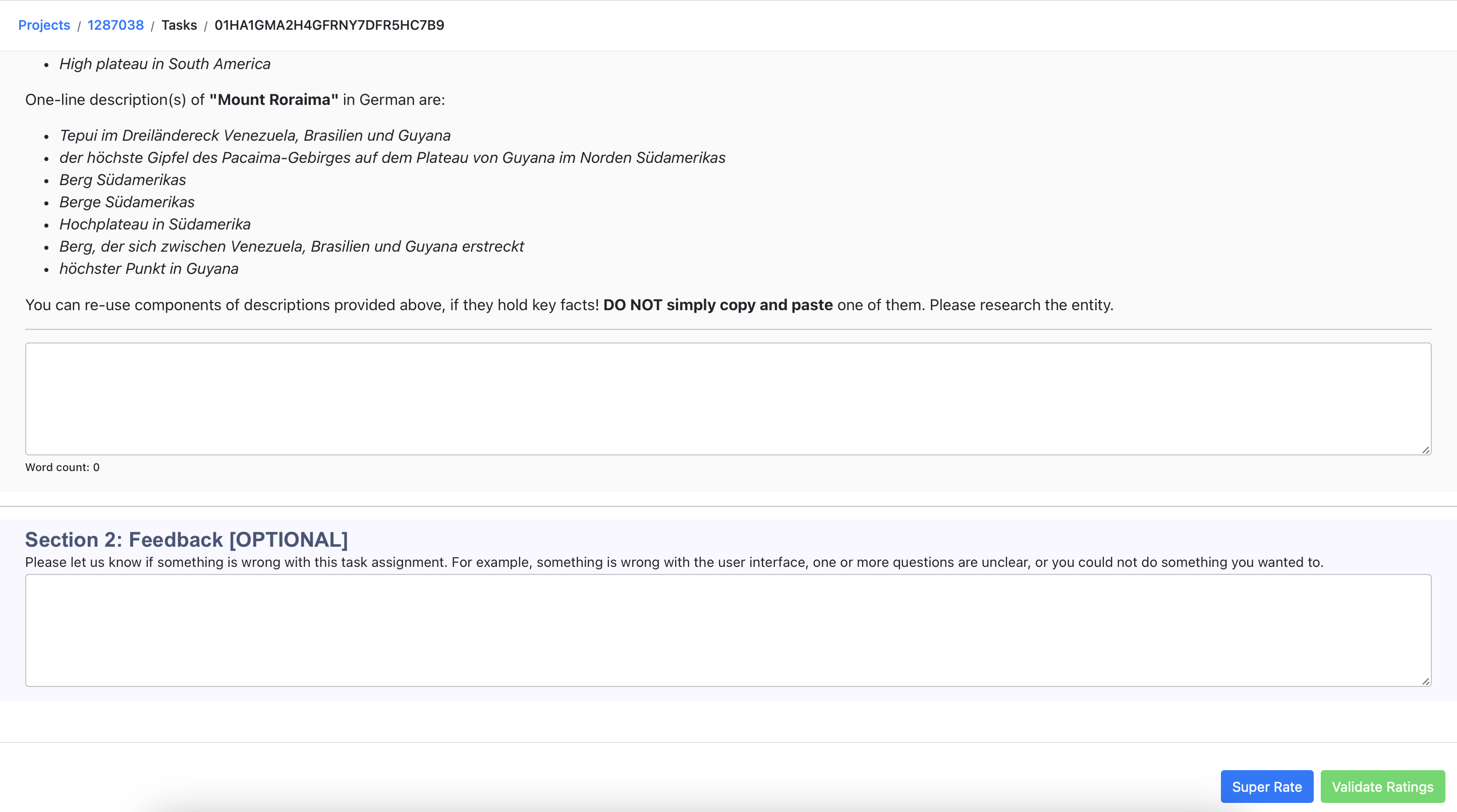}
  \caption{UI used for the annotation task: the annotator provied the description in a text box. A warning message was prompted if the token length of the description was too short or too long.}
  \label{fig:ui-third-part}
\end{figure*}

\subsubsection{Validate entity descriptions.}
The objective of this annotation step was to validate the quality of the descriptions in the target language provided by the human annotators.

First, given an entity, the human annotator was provided with corresponding information (i.e., entity names/aliases, Wikipedia pages etc.) as done in the previous task. In addition, the description requirements were detailed (in-depth guidelines provided in a different document).

Then, they were prompted to analyze the corresponding description for the entity in the target language with a series of questions. The questions were reformulated from description requirements, to verify the presented description in the target language followed the requested format. If the annotator negatively responded to any of the presented questions, they were prompted to edit the description to satisfy the requirements. If the initial description meets the requirements, the originally provided description was sustained.

\begin{figure*}[t]
  \centering
  \includegraphics[width=\textwidth]{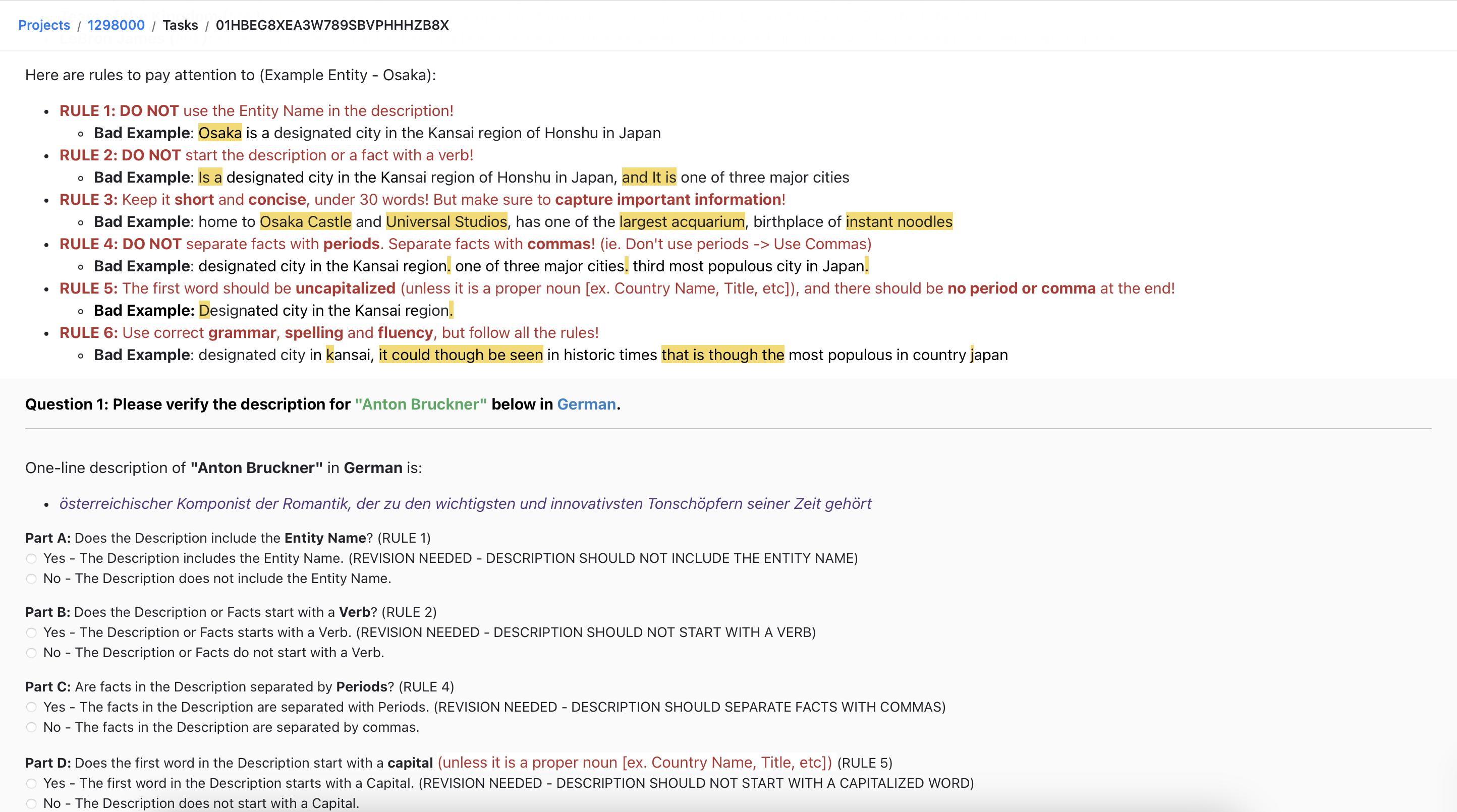}
  \caption{UI used for the annotation task: annotators were required to examine the description in the target language, and answer a series of questions that reflected the description requirements.}
  \label{fig:ui-third-part}
\end{figure*}

\begin{figure*}[t]
  \centering
  \includegraphics[width=\textwidth]{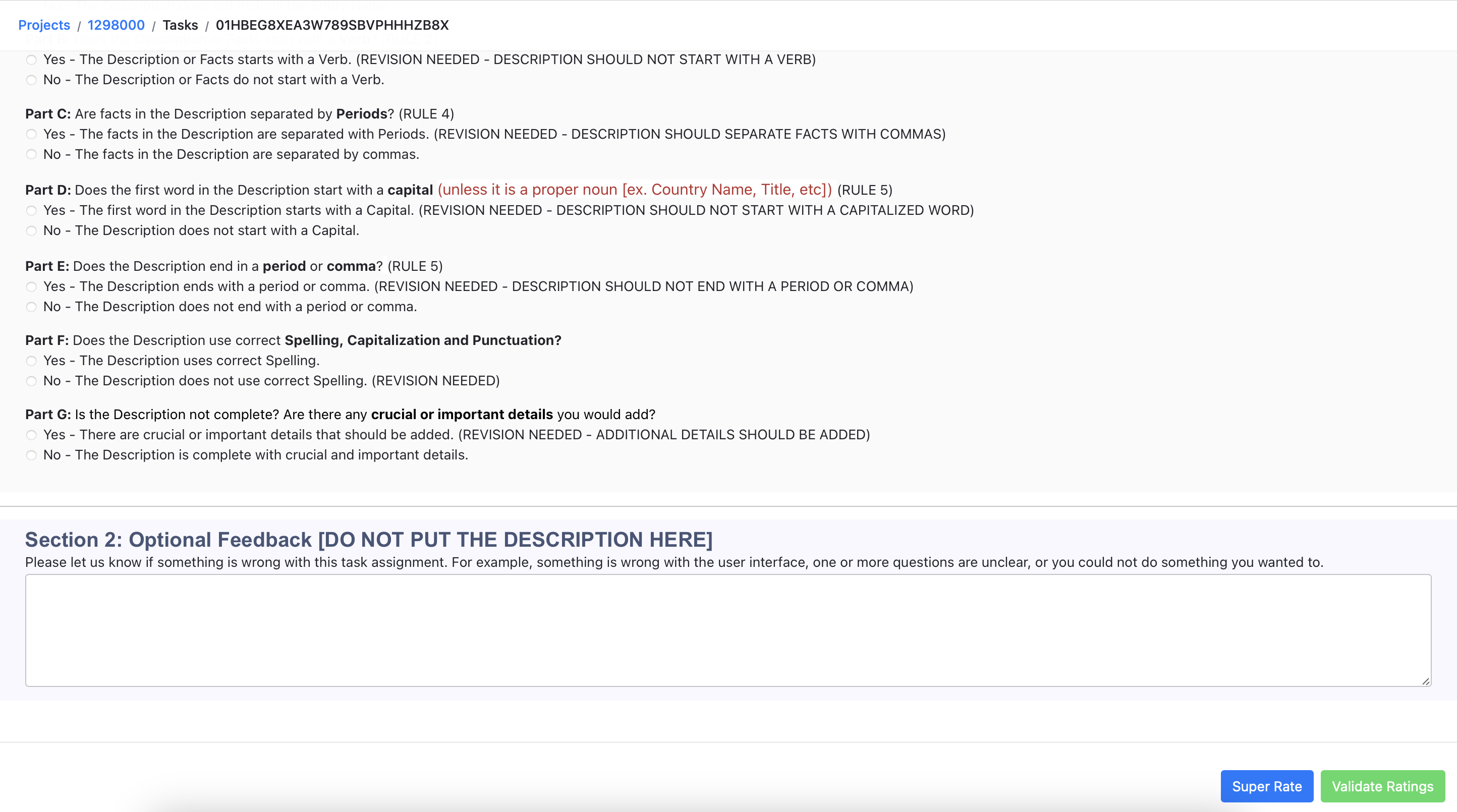}
  \caption{UI used for the annotation task: annotators were prompted to correct the description by rewriting it, if they negatively answer the series of questions provided.}
  \label{fig:ui-third-part}
\end{figure*}

\subsection{Quality assurance and inter-annotator agreement.}
We follow the annotation guidelines by \citeauthor{conia-etal-2023-increasing} to get high quality annotation results for the datasets. 
As the entire annotation procedure was done in a crowd-source platform outside of our organization, we will only disclose necessary information to protect annotators' privacy and meanwhile ensure the quality of \benchmark{}.
Each item in \benchmark{} is annotated by 3 annotators. All annotators are native speakers of the language they annotate, and they are fluent in English too. 
We follow the evaluation protocol proposed by \citeauthor{conia-etal-2023-increasing}. Inter-annotator agreement on head, torso and tail entity names resembles what is reported in \citeauthor{conia-etal-2023-increasing}, reflected by pairwise Cohen’s Kappa around 0.8 and Krippendorff’s alpha around 0.95, which shows strong agreement across annotators.


\section{Short Description Evaluation}

\begin{table*}[t]
  \centering
  \adjustbox{max width=0.9\textwidth}
  {
    \begin{tabular}{llcccccccccc}
      \toprule
      \multirow{6}{*}{\rotatebox[origin=c]{90}{\textit{descriptions}}} & \textbf{BLEU Score}                                                    & \textbf{\#Params}     & \textbf{AR}  & \textbf{DE}  & \textbf{ES}  & \textbf{FR}  & \textbf{IT}  & \textbf{JA}   & \textbf{KO}  & \textbf{ZH}  & \textbf{Avg} \\
      \cmidrule(l{3pt}r{3pt}){2-12}
                                                                       & NLLB-200$_{\small\ \texttt{EN} \rightarrow \texttt{XX}}$ $\rightarrow$ & 0.6B                  & 2            & 3.4          & 7            & 4.8          & 4.6          & 1.5           & 4.1          & 5.9          & 4.2          \\
                                                                       & GPT-3.5                                                                & 175B                  & \textbf{2.8} & \textbf{4.2} & 7.2          & 4.9          & 5.5          & \textbf{3.9 } & \textbf{4.3} & \textbf{9.8} & \textbf{5.3} \\
      \cmidrule(l{3pt}r{3pt}){2-12}
                                                                       & $\mathsf{TRICK_{KGE}}$                                                 & \multirow{3}{*}{0.6B} & 2.2          & 3.4          & \textbf{8.0} & \textbf{5.1} & \textbf{5.9} & 3.8           & 2.5          & 4.9          & 4.5          \\
                                                                       & $\mathsf{TRICK_{50\%KGC+KGE}}$                                         &                       & 1.4          & 2.1          & 5.8          & 2.8          & 4.1          & 2.3           & 1.6          & 2.8          & 2.9          \\
                                                                       & $\mathsf{TRICK_{KGC+KGE}}$                                             &                       & 1            & 0.8          & 2.6          & 1.3          & 2.1          & 2             & 1.5          & 2.2          & 1.7          \\

      \bottomrule
    \end{tabular}
  }
  \caption{BLEU score for entity short description evaluation}
  \label{tab:kge_bleu}
\end{table*}

As is shown in Table \tabref{tab:kge_bleu}, we calculate the BLEU score for every baseline and our method. However, the BLEU score for all languages and all baselines are under 10, which suggests that the translated text from English can hardly relate to ground truth in target languages. This phenomenon could suggest that BLEU is not a proper metric for entity short description evaluation, as (i) Short description for the same entity in different languages are not directly translatable. (ii) A large amount of short descriptions are less than 4 tokens (e.g. \textit{Politician}), which could bias the judgement of BLEU when calculating the weighted average.

\section{KG-TRICK Training Configurations}
For both KGC and KGE tasks including all variants of KG-TRICK models, we initiate from mBART-50-large and set maximum of training epochs to 6, aligning with KG-T5’s implementation. We set batch size to be 48, Learning Rate to be 8e-4 with Adam optimizer and used a scheduler (i.e. transformers.get\_inverse\_sqrt\_schedule) during training.

\section{Balancing the training data for KGC and KGE}\label{sec:kgc_kge_ratio}
In this section, we provide more details on how balancing the training data between KGC and KGE tasks can impact the performance of the two tasks.
Indeed, the training datasets available for the two tasks are not balanced: the KGC dataset contains 150 million records generated multilingually from 20 million triplets, while the KGE dataset contains around 16 million records generated multilingually from 5 million entities.
Therefore, we investigate the impact of mixing different proportions of the two datasets on the performance of the two tasks.
More specifically, we investigate different proportions of the KGC and KGE datasets, ranging from 0\% to 100\% of the KGC dataset, and evaluate the performance of the two tasks on WikiKGE-10++.
The results are reported in Table~\ref{tab:kgc_ratio}.
We can observe that the best performance on KGC is achieved when the full KGC dataset is used, which suggests that the KGC task is more difficult than the KGE task.
On the other hand, the best performance on KGE is achieved when up to 50\% of the KGC dataset is used.
Therefore, the best compromise between the data mixing proportion for the two tasks is to use 50\% of the KGC dataset.

\begin{table}[ht]
  \centering
  \adjustbox{max width=\linewidth}{
    \begin{tabular}{lcccc}
      \toprule
            & \multicolumn{2}{c}{\textit{KGE}} & \multicolumn{2}{c}{\textit{KGC}}                                 \\
      \cmidrule(l{3pt}r{3pt}){2-3} \cmidrule(l{3pt}r{3pt}){4-5}
      KGC\% & Precision                        & Coverage                         & MRR           & hit@1         \\
      \midrule
      0\%   & 51.5                             & 30.9                             & -             & 30.4          \\
      1\%   & \textbf{52.4}                    & 30.4                             & 32.7          & 32.1          \\
      10\%  & \textbf{52.4}                    & 30.6                             & 34.4          & 31.7          \\
      20\%  & 51.7                             & 28.8                             & 34            & 32.8          \\
      50\%  & 52.1                             & \textbf{31.5}                    & 35.2          & 33            \\
      full  & 52.2                             & 29.7                             & \textbf{38.8} & \textbf{36.6} \\
      \bottomrule
    \end{tabular}
  }
  \caption{Investigation on the different mixing proportion between KGC and KGE training data, and their impact on KGC and KGE tasks performance}
  \label{tab:kgc_ratio}
\end{table}

\end{document}